\documentclass[letter, 10 pt, conference]{ieeeconf}  

\IEEEoverridecommandlockouts                              

\usepackage{fancyhdr}
\usepackage{xcolor}
\usepackage[some, placement=top]{background}

\usepackage{amsmath}
\usepackage{amssymb}
\usepackage{mathtools}
\usepackage{graphicx}
\usepackage{float}
\usepackage{array}
\usepackage{tikz}
\usepackage{listings}
\usepackage{hyperref}
\usepackage{graphicx}
\usepackage{cite}
\usepackage{dsfont}
\usepackage{mathtools,etoolbox}
\usepackage[ruled, linesnumbered]{algorithm2e}
\usepackage[none]{hyphenat}
\usepackage[utf8]{inputenc}
\usepackage[english]{babel}
\usepackage[T1]{fontenc}

\usepackage{amsthm}
\usepackage{xfrac}
\usepackage{nicefrac}
\usepackage{booktabs}
\usepackage{flushend}
\usepackage[switch, pagewise]{lineno}
\usepackage{mathrsfs}
\DeclareMathAlphabet{\mathpzc}{OT1}{pzc}{m}{it}
\usepackage{multirow}
\usepackage{multicol}
\usepackage{bbm}
\usepackage{soul}
\usepackage{xfrac}
\usepackage{balance}
\usepackage{titlesec}
\usepackage[final]{pdfpages}

\usepackage{amsmath}
\usepackage{subfigure}
\usepackage{afterpage}
\usepackage{resizegather}
\usepackage{nicematrix,tikz,booktabs}

\DeclareMathOperator*{\argmin}{argmin} 

\hypersetup{colorlinks=true,urlcolor=blue,citecolor=red}

\title{\LARGE \bf \textit{MinNav}: Minimalist Navigation For Active Tiny Aerial Robots}

\author{Aniket Patil$^{1}$, Mandeep Singh$^{1}$, Uday Girish Maradana$^{1}$, Nitin J. Sanket$^{1, *}$
\thanks {$^1$Perception and Autonomous Robotics (PeAR) Group, Robotics Engineering Department, Worcester Polytechnic Institute.
{\textit{$^*$Corresponding author: Nitin J. Sanket} (\texttt{nitin@wpi.edu}).}}}

\hypersetup{
    colorlinks=true,
    linkcolor=blue,
    filecolor=magenta,      
    urlcolor=cyan,
}

\backgroundsetup{
    placement=top,
    position={8.0cm,1.75cm}, 
    scale=1,
    angle=0,
    opacity=1.0, 
    color=gray,
    contents={\parbox{16cm}{\normalsize \centering This paper has been accepted for publication at the IEEE International\\Conference on Robotics and Automation (ICRA), Vienna, 2026. \textcopyright~IEEE}}
}

\begin{document}

\makeatletter
\g@addto@macro\@maketitle{%
\begin{figure}[H]
   \setlength{\linewidth}{\textwidth}
   \setlength{\hsize}{\textwidth}
    \centering
    \includegraphics[width=1\textwidth]{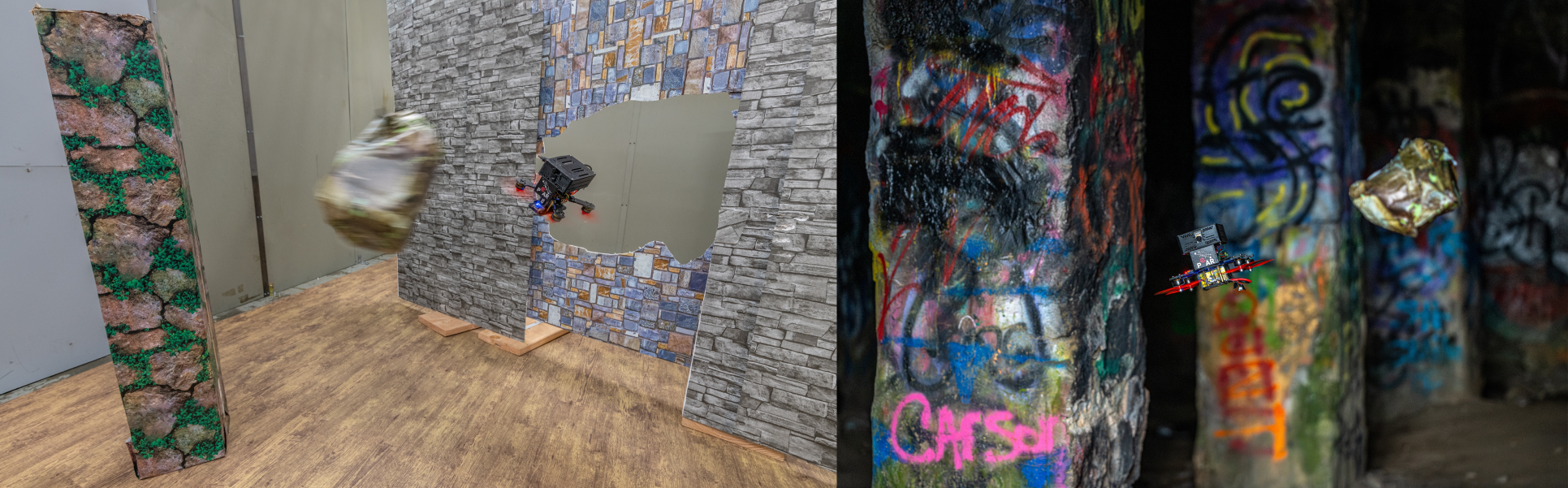}
    \caption{\textit{MinNav} handles navigation in unstructured and wild scenes including static obstacles, dynamic obstacles and unknown shaped gaps without any prior knowledge of location or scene ordering. All this is performed using a monocular camera and an active strategy using only onboard computation and sensing. \textit{All the images in this paper are best viewed in color on a computer screen at 200\% zoom.\\[-40pt]}}
    \label{fig:Banner}
\end{figure}
}
\makeatother

\BgThispage

\maketitle
\setcounter{figure}{1}

\begin{abstract}
Navigation using a monocular camera is pivotal for autonomous operation on tiny aerial robots due to their perfect balance of versatility, cost and accuracy. In this paper, we introduce \textit{MinNav}, a navigation stack based on optical flow and its uncertainty to fly through a scene with static and dynamic obstacles and unknown-shaped gaps without any prior knowledge of the scene components and/or their locations/ordering. We further improve success rate by using the activeness of the robot to move around in an exploratory way to find obstacles and navigate. We successfully evaluate and demonstrate the proposed approach in many real-world experiments in various environments with static and dynamic obstacles and unknown-shaped gaps with an overall success rate of 70\%. To the best of our knowledge, this is the first solution to tackle all the aforementioned navigation cases without prior knowledge using a monocular camera. Our approach is on par in performance with depth based methods with factors of magnitude less computation required and can readily run onboard tiny aerial robots.   
\end{abstract}



\section*{Supplementary Material}
The accompanying video, supplementary material, code and dataset can be found at \url{https://pear.wpi.edu/research/minnav.html}.

\section{Introduction}

Tiny aerial robots have risen in popularity in the last decade due to their utility in humanitarian applications such as plant pollination \cite{mazinani2023design, sanket2021active}, search and rescue \cite{SearchAndRescue, mohta2016quadcloud, delmerico2019current} among others \cite{Inspection}. This rapid rise has been accelerated by the development of better computing hardware at lower power and size\cite{raju2024edgeflownet, sanket2021prgflow}. While these robots offer advantages in agility, safety, and scalability, their autonomy remains limited by sensor and computational constraints. The use of Optical Flow rather than a 3D reconstruction of the scene has emerged as a more efficient representation, as remarked by experts in computer vision and robotics \cite{duchon1995ecological, souhila2007optical, horn1980determining, ActiveVision}. Using optical flow for navigation is minimalist in operation and works on image pixels rather than metric units for navigating a robot through a scene \cite{sanket2018gapflyt, sanket2021active, sanket2023ajna, bouwmeester2023nanoflownet}. Variants of optical flow are speculated to be used in nature, where bees perform peering and other maneuvers to dodge obstacles or go through a gap using functions of optical flow \cite{burnett2020wind}. Computational algorithms inspired by nature have been used for performing various navigational tasks such as flying through gaps \cite{sanket2018gapflyt}, dodging static and dynamic obstacles \cite{sanket2020evdodgenet}, landing \cite{dupeyroux2021neuromorphic} and so on. As some of the prior works point out, using only optical flow for navigation has a few problems: it has a dead spot near the focus of expansion where the flow magnitudes are low \cite{stevens2018vision}, flow can be ill-conditioned near color flat regions, large changes in illumination, around object boundaries and so on \cite{shah2021traditional, sanket2023ajna}. Recently, works have proposed to utilize uncertainty in flow predictions \cite{sanket2023ajna} as a representation for navigation in various scenes, but this requires prior knowledge of the scene setup or the kind of scene that might be encountered (static or dynamic obstacles). 

To this end, we present \textit{MinNav}, which utilizes optical flow and its uncertainty to navigate through a scene that has static obstacles, dynamic obstacles and unknown gaps without any prior knowledge of the ordering of the scene components with only onboard sensing and computation. We utilize a neural network to obtain optical flow and uncertainty prediction onboard a tiny aerial robot. We further use the mathematical properties of our formulation to intuit how we can classify among various scene motifs (types of scene components). This classification enables a first-of-its-kind unified framework that jointly addresses all three motifs to generate efficient navigation commands, surpassing prior works that treat flow and uncertainty in isolation.

\subsection{Problem Formulation and Contributions}
\label{subsec:contrib}
A quadrotor is present in a scene that can have static obstacles, a static unknown-shaped gap and/or dynamic obstacles (called scene motifs). The quadrotor has no prior knowledge of which scene motifs are present and/or their ordering, size, shape or location. The robot is tasked to perform goal-directed navigation (such as heading north) using only a front-facing monocular camera. A down-facing optical flow sensor, laser altimeter, and inertial measurement unit are employed strictly for low-level attitude control, velocity estimation, and altitude regulation. All sensing and processing is done on board without the use of motion capture or GPS. 
We present an active approach to fly toward the free space while avoiding unknown static and dynamic obstacles. The core concept is built around utilizing optical flow and its uncertainty to find free space regions for navigation.
A summary of our contributions are (Fig. \ref{fig:Banner}):
\begin{itemize}
    \item We propose a generalized navigation stack for a monocular camera-equipped robot to navigate through a scene with unknown static obstacles, gaps and dynamic obstacles. To the best of our knowledge, this is the first work that handles all such scenarios together ($\S$\ref{sec:nav}).
    \item We present a lightweight multi-scale optical flow and uncertainty network that can run onboard embedded computers (Fig. \ref{fig:arch}, $\S$\ref{subsec:network}).
    \item We propose an active control policy that combines exploratory motion and moving towards the free space in one formulation ($\S$\ref{sec:control}). 
    \item We evaluate and demonstrate the proposed approach on a real quadrotor with onboard perception and computation in many real-world experiments and simulation experiments ($\S$\ref{sec:Expts}). To benefit the community, we will release the source code and simulation environments used in this work.
\end{itemize}

\subsection{Related Work}
\label{subsec:related}
There are two main trains of thought when it comes to navigation through cluttered spaces using aerial robots.

\subsubsection{Depth-based/Structured Navigation}
In these classes of methods, structure or depth is either directly perceived by a sensor(s) or by reconstructing the environment. The most common way for achieving such navigation is by the use of commonly found LIDAR sensors \cite{zhang2017online, ren2025safety} for larger robots and RGBD sensors for smaller robots\cite{zhou2022swarm, kulkarni2023semantically}. Classically, a 3D representation is used on which paths and trajectories are planned and executed \cite{gao2019flying, zhou2022swarm, liu2016high}, but recent studies have shown that combining or replacing perception, planning and control using deep learning is more efficient and often more performant \cite{hanover2023autonomous, song2023reaching, kaufmann2023champion, eschmann2023learning, ferede2024end} either due to better policies learned or finding better policies through reinforcement learning. Although structured approaches seem like the gold-standard solution for navigation, there could be often cases when estimating depth using an active sensor might not be allowed or possible (such as direct sunlight or dusty conditions), or the sensor (or computation) might be too heavy to carry on a tiny robot. This necessitates a structureless approach to navigation. One can also imagine structureless approaches based on a monocular camera always running in the background and taking over during a sensor failure or low battery condition of a structured approach for larger robots.   


\subsubsection{Structureless Navigation}
Structureless Navigation refers to methods that perform navigation without explicitly computing metric or relative depth (commonly called structure in computer vision literature). In these methods, either a function of depth is perceived that is directly used for control \cite{izzo2012landing, gapflyt}, or lately, in tiny robots, motion is used to obtain additional information cues to infer about functions of depth \cite{gapflyt, bouwmeester2023nanoflownet}. One such formulation utilizes optical flow (motion of pixels) to infer the ordinality of depth using the parallax effect (closer objects move more than farther objects). Newer works have also shown the efficacy of using optical flow uncertainty or occlusion masks to perform navigational tasks \cite{sanket2023ajna}. Although, in general, structureless approaches are sub-optimal in terms of path length as compared to structured approaches, they often use lesser computation power and lower-quality sensors making them perfectly suited for tiny robots.

In both cases, dynamic obstacles are handled very differently. In structured approaches, the simplest way is to treat the dynamic obstacle as a static obstacle and dodge it when it is closer than some distance. A better way is to model the 3D motion and predict the movement to dodge more efficiently. Structureless approaches either use a motion detection sensor such as an event camera to find dynamic obstacles efficiently \cite{sanket2020evdodgenet} or optical flow coupled to Epipolar geometric constraints\cite{zhong2019unsupervised} or optical flow inlier probability (or uncertainty) to find dynamic obstacles\cite{sanket2023ajna, ranjan2019competitive}. In the latter case, the obstacles are dodged in a best-effort manner.

To the best of our knowledge, no prior work handles all cases of navigation: dodging dynamic and static obstacles and flight through unknown gaps using a single structureless formulation without prior information. To this end, we present a novel method based on optical flow and its uncertainty to achieve navigation in a scene with static and dynamic obstacles and unknown gaps without any prior knowledge using only onboard sensing and computation using a monocular camera without reconstructing the scene.

\subsection{Organization of the paper}
\label{subsec:organization}
The paper is structured into perception and control modules. We present our perception solution in $\S$\ref{sec:nav} where we explain how we distinguish between different scene cases (static obstacles, dynamic obstacles and gaps). After which we talk about our active perception-based control policy to enhance our perception stack while navigating towards the goal direction in $\S$\ref{sec:control}. Experiments and analysis for both real-world and simulation settings are detailed in $\S$\ref{sec:Expts} and finally, we conclude the paper in $\S$\ref{sec:Conc} with parting thoughts on future work.

\section{Scene Motifs For Visual Navigation}
\label{sec:nav}
When navigating in the wild, an aerial robot can encounter various scene components such as static obstacles, dynamic obstacles and free space. In this work, we will call a collection of such scene components as \textit{scene motifs}. Commonly, five scene motifs are found in both natural and human-made environments: (a) Free space with static obstacles, (b) Free space with dynamic obstacles, (c) Static Gaps, (d) Dynamic Gaps, (e) Dead-end, i.e., no free space. In this work, we propose to navigate through a scene in the first three scenarios where we do not have any prior knowledge of the scene statistics or geometry only using an on-board monocular camera with on-board computation. To enable such a general navigation stack without explicit 3D scene reconstruction, we utilize motion information between image frames in the form of optical flow and its uncertainty. Before we explain our approach further, we will explain a few preliminaries required. 

\subsection{Preliminaries}
\label{subsec:prelim}
Let two image frames captured at $t$ and $t+\delta t$ be denoted as $\mathcal{I}_t$ and $\mathcal{I}_{t+\delta t}$ respectively. Further, let the 3D linear and angular velocities of the camera in this time be $V=\begin{bmatrix} V_x & V_y & V_z\end{bmatrix}^T$, $\Omega=\begin{bmatrix} \Omega_x & \Omega_y & \Omega_z\end{bmatrix}^T$. Optical flow at a pixel  $\mathbf{x}=\begin{bmatrix} x & y\end{bmatrix}^T$ is defined as the apparent image velocity of the corresponding 3D world point and is given by

\begin{equation}
    \mathbf{\dot{p}}_{\mathbf{x}} = \dfrac{1}{Z_\mathbf{x}}
    \begin{bmatrix}
    x V_z - V_x \\
    y V_z - V_y \\
    \end{bmatrix} + \begin{bmatrix}
    xy & -(1+x^2) & y \\
    (1+y^2) & -xy & -x \\
    \end{bmatrix} \Omega
\end{equation}


Here, $Z_\mathbf{x}$ denotes the depth at a pixel $\mathbf{x}$ which is commonly called the \textit{structure}. Estimating 3D structure or reconstructing the scene is computationally expensive and slow for tiny robots. To navigate through a scene, we wish to distinguish between different motifs which would be simple if the structure was known.  In this work, we utilize optical flow and its uncertainty to distinguish between various motifs and take the appropriate control action for navigation. Intuitively, closer objects (obstacles) produce high optical flow (particularly translational flow) by virtue of the low value of $Z_\mathbf{x}$. One might ponder, this is very simple then, why not just use the magnitude of optical flow for control action? There are two main issues (a) the rotational flow which is independent of the scene can overshadow the translational flow during fast rotational motion commonly found in aerial robots, and (b) the translational flow has a low magnitude near the Focus Of Expansion (FOE) which cannot be used for control action due to inherent high uncertainty and high noise sensitivity. A common solution for the first problem is either by using a mechanical or an electronic gimbal\cite{stevens2018vision}. For the second problem, past works \cite{stevens2018vision} avoid utilizing the circular area around the FOE for control. To tackle both issues, we propose to utilize the active nature of the robot, i.e., we can make controlled exploratory motions to reason about the scene to take a navigation action using the philosophy of Active Perception\cite{ActiveVision, BajcsyActive}. By maintaining small rotational movements in our exploratory motion (cf. $\S$\ref{sec:control}.), we obtain both low rotational flow and maintain FOE outside the image plane. Our perception stack for each motif is explained in the subsequent subsections.

\subsection{Static Obstacles and Unknown Gaps}
Since gaps are special cases of free space with obstacles, we treat these two motifs as the same entity. Technically, gaps are enclosed free spaces surrounded by obstacles. But for all practical purposes, we need to fly to the free space avoiding obstacles. In the case of static obstacles, if we have `tackled' the FOE and high rotation issues as we discussed previously, then we can simply look for the largest contour (largest free space) of low flow magnitude and fly towards it. Contrary to previous works, our approach trades off perception complexity with planning and control complexity. Our perception stack is very simple due to carefully controlled exploratory active movements. The price to pay here is in terms of exploratory control action which is generally negligible for small agents as we see in nature \cite{chittka2023mind, ravi2019gap}.

\subsection{Dynamic Obstacles}
Dynamic obstacles are defined as obstacles whose movement is independent of one's movement (ego-motion). This means that they will generate an optical flow field with a different FOE to the robot's own movement. Prior works have clustered optical flow based on FOE\cite{zhong2019unsupervised}, but this is expensive because of the number of dynamic obstacles and other statistics are not known. One might ponder, can we not just look for high optical flow regions for dynamic obstacles? Conceptually, we want to dodge the obstacle only if it is coming toward us, in which case the optical flow magnitudes will be higher as compared to a static obstacle at the same depth $Z_\mathbf{x}$ because the relative velocities ${V, \Omega}$ will be higher. But this is not as simple as it sounds when we have both static and dynamic obstacles in the scene with similar optical flow magnitudes. Furthermore, to dodge dynamic obstacles, one has to dodge the `future' projection not just the current detection since the object is moving. To this end, we will utilize optical flow uncertainty\cite{sanket2023ajna} to gain additional cues which we will use to distinguish static from dynamic obstacles with large flow magnitudes. In this work, since we do not know when we might encounter a dynamic obstacle, we utilize a combination of high flow magnitude and high uncertainty to detect dynamic obstacles. High uncertainty is encountered for dynamic obstacles due to large amounts of occlusion presented when objects move.

\begin{figure*}[t!]
    \centering
    \includegraphics[width=0.95\textwidth]{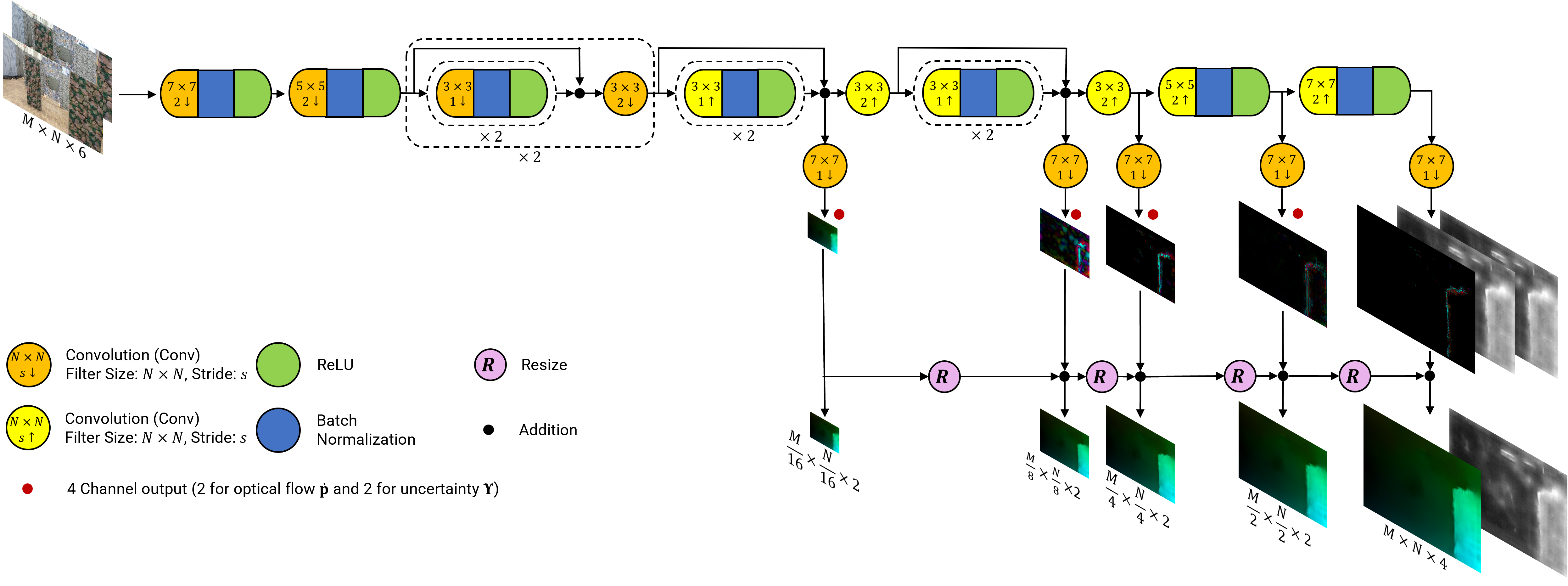}
    \caption{Lightweight multi-scale pyramidal neural network architecture used to predict optical flow $\mathbf{\dot{p}}$ and its uncertainty $\mathbf{\Upsilon}$ used in \textit{MinNav}.}
    \label{fig:arch}
\end{figure*}

\subsection{Network Architecture and Training Details}
\label{subsec:network}

Our network architecture is inspired from previous works \cite{sanket2023ajna, sanket2021evpropnet, singh2021nudgeseg}. We perform the following modifications:  (a) Multiscale pyramidal architecture for predicting optical flow at different levels in a coarse to fine manner, (b) Our multiscale architecture differs from the traditional approaches, such as in \cite{ranjan2016opticalflowestimationusing}, by only predicting incremental optical flow for an efficient design. For implementation we use five levels ($L=5$) and the architecture is shown in Fig. \ref{fig:arch}. Our network is a fully convolutional architecture and can take any arbitrary input size of $M\times N$ and output the same size. Specifically, the input is a stack of temporal images of size $M \times N \times 6$ and the output is optical flow $\dot{\mathbf{p}}$ and its uncertainty $\Upsilon$ with a final output size of $M \times N \times 4$ (including $x$ and $y$ directions).    



Our network was trained for 400 epochs on the FlyingChairs2 (FC2) dataset \cite{DFIB15,ISKB18} and then fine-tuned for another 50 epochs on the FlyingThings3D (FT3D) dataset \cite{MIFDB16} for better generalization to the real world. Our \textit{MinNav} network has 2.8M parameters which is small enough to run on-board the Jetson Orin Nano with a 68.2$ms$ forward inference time without any optimizations such as TensorRT on $1 \times 640 \times 480 \times 6 px.$ resolution (24.6$ms$ for $1 \times 320 \times 240 \times 6 px.$). The network is trained using a custom multiscale loss function (Eqs. \ref{eq:LossFunc1}, \ref{eq:LossFunc2}) with ADAM optimizer and a learning rate of $10^{-4}$ for the first 400 epochs (FC2) and then $10^{-5}$ for the last 50 epochs (FT3D) with a mini-batch size of 32.   


\begin{gather}
\label{eq:LossFunc1}
\mathcal{L} =  \argmin_{\mathbf{\dot{p}_x}, \mathbf{\Upsilon_x}} \sum_{l=1}^L\left (\mathbb{E}_{\mathbf{x}}\left( \frac{\Vert \mathbf{\dot{p}}_{\mathbf{x}, l} - \mathcal{R}_{L}^l \left(\mathbf{\dot{q}_x}\right) \Vert_1}{\log \left( 1 + e^{\left( \mathbf{\Upsilon}_{l,\mathbf{x}} + \epsilon \right)}\right)} +   \log \left( 1 + e^{\mathbf{\Upsilon}_{l,\mathbf{x}}}\right) \right)  \right)
\end{gather}
\begin{equation}
\label{eq:LossFunc2}
    \mathbf{\dot{p}_x}=\sum_{l=2}^{L} \left( \mathcal{R}_{l-1}^l \left( \mathbf{\dot{p}}_{\mathbf{x},l-1} \right) + \Delta \mathbf{\dot{p}}_{\mathbf{x},l} \right); \quad \mathbf{\dot{p}}_{ \mathbf{x},1} = \Delta  \mathbf{\dot{p}}_{\mathbf{x},1}
\end{equation}
Here, $\mathbf{\dot{p}_x}, \mathbf{\dot{q}_x}, \mathbf{\Upsilon_x}$ are the predicted optical flow, ground truth optical flow and predicted uncertainty respectively.  $ \Delta \mathbf{\dot{p}}_{l, \mathbf{x}}$ denotes the incremental flow prediction at level l and $\mathcal{R}_{l-1}^l \left( \mathbf{\dot{p}}_{l-1, \mathbf{x}} \right)$ represents a differentiable bilinear resizing function that resizes optical flow from size at level $l-1$ to $l$ and $\epsilon$ is a small constant for numerical stability (we use $10^{-3}$ in our experiments). Finally, $L$ is the number of levels which is set to 5 in our experiments. 
At the heart of the perception stack is the exploratory motion that is imbibed in the control stack as described next.

\section{Active Perception Based Control Strategy}
\label{sec:control}
\begin{figure}[t!]
    \centering
    \includegraphics[width = 0.9\linewidth]{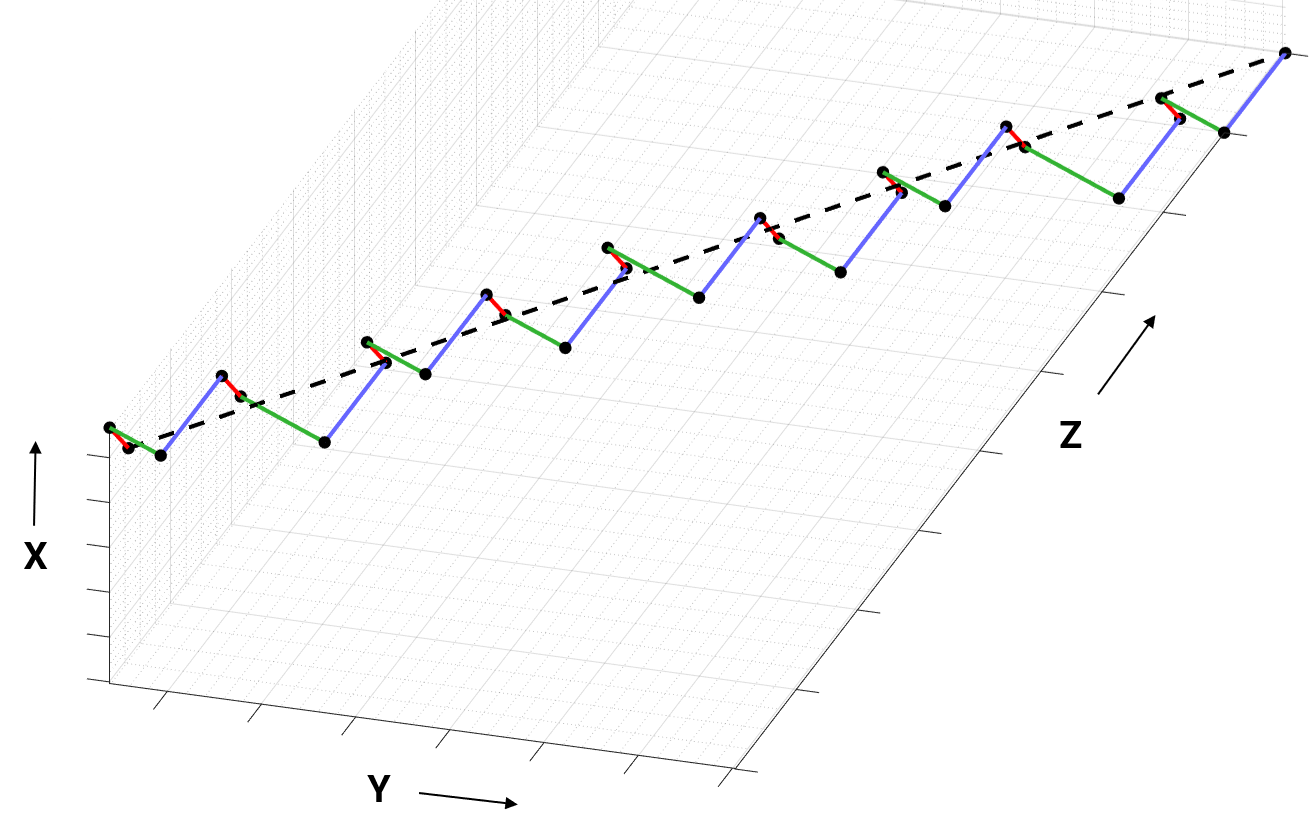}
    \caption{A toy example showing how our control strategy will ``wander'' around the perfect line the robot would have followed if metric depth was available (dashed black line, time increasing with increasing $Z$). The \textcolor[rgb]{0.8,0,0}{red} parts show tiny active exploration motion, \textcolor[rgb]{0,0.8,0}{green} parts show free space alignment motion  and \textcolor[rgb]{0,0,0.8}{blue} parts show the forward $Z$ motion. }
    \label{fig:drone_control_algo}
\end{figure}

The core idea behind our control strategy involves three key steps: (a) a diagonal exploratory motion (in X and Y direction), (b) a movement towards free space (in X and Y direction) and (c) a movement forward towards the global goal direction (Z axis). The concept of exploratory motion is fundamentally inspired by concepts of Active Perception \cite{ActiveVision, BajcsyActive, sanket2018gapflyt, bouwmeester2023nanoflownet} and Honey Bee Peering \cite{chittka2023mind, ravi2019gap, ravi2022bumblebees} as explained next. See Fig. \ref{fig:drone_control_algo} for a toy example.

\subsection{Static Obstacle Avoidance And Flight Through Unknown Gaps}
We utilize the above strategy to avoid static obstacles and also fly through static unknown gaps by iteratively moving toward the free space region.

\textit{Active Exploratory Motion:}
As we mentioned before, the goal of the exploratory motion is to avoid the issues that arise around FOE. To this end, we employ a diagonal motion in the 3D X-Y plane. The exploratory motion is implemented using an open-loop position controller with a pre-chosen period and velocity profile for the diagonal motion. Let $\tau$ be the period of diagonal motion, $P^E$ be the open-loop exploratory positional command, $V=\begin{bmatrix} V_x & V_y & V_z\end{bmatrix}^T$ be the chosen velocity, then the commanded motion is given by:
\begin{equation}
{P^E} = \begin{bmatrix}
    P_x^E &
    P_y^E &
    P_z^E 
\end{bmatrix}^T = \begin{bmatrix}
    {V_x}^E \tau &
    {V_y}^E  \tau &
    0
\end{bmatrix}^T
\end{equation}

\begin{figure*}[t!]
\centerline{\includegraphics[width=1.0\textwidth]{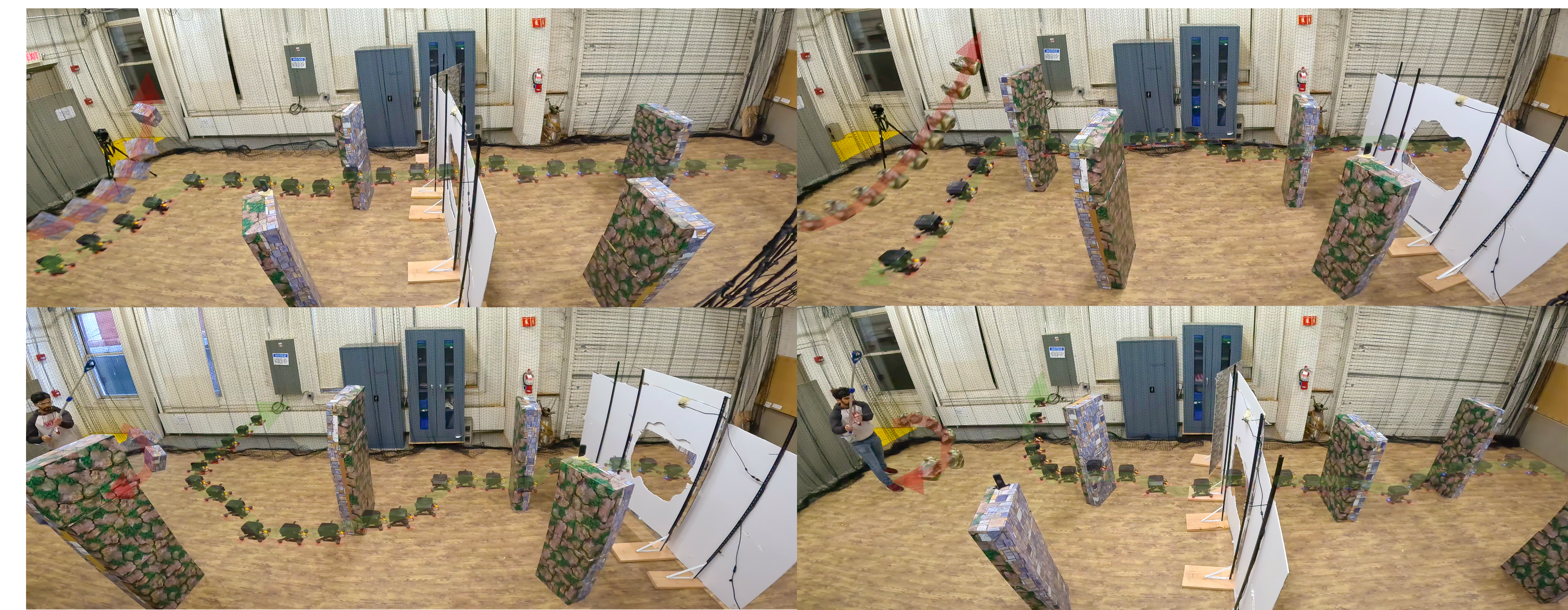}}
\caption{ Sequence of images of quadrotor navigating through different scenarios. \textcolor[RGB]{0, 200, 0}{Green} and \textcolor[RGB]{200, 0, 0}{red} arrow shows the path of the robot and the dynamic obstacle respectively.}
\label{fig:experiments}
\end{figure*}

Note that, we alternate the sign of both $V_x^E$ and $V_y^E$ at every iteration to have minimal drift.

\textit{Alignment Movement Towards Free Space:} Since our goal is to keep moving towards free space, once we obtain image frames after exploration, we take a step towards the free space using a simple velocity controller. Let $\mathbf{x}_0=\begin{bmatrix} x_0 & y_0 & 1\end{bmatrix}^T$ and $\mathbf{x}_\mathcal{F}=\begin{bmatrix} x_\mathcal{F} & y_\mathcal{F} & 1\end{bmatrix}^T$ be the center of the image and the center of the free space region on the image plane, respectively. We implement a alignment open-loop incremental position command $P^\mathcal{F}$ using a velocity controller and a fixed time period as before. 
\begin{equation}
{P^\mathcal{F}} = \begin{bmatrix}
    P_x^\mathcal{F} &
    P_y^\mathcal{F} &
    P_z^\mathcal{F} 
\end{bmatrix}^T = \begin{bmatrix}
     {K_{p, x}}\left(x_0 - x_\mathcal{F}\right) \tau &
    {K_{p, y}}\left(y_0 - y_\mathcal{F}\right) \tau &
    0 
\end{bmatrix}^T
\end{equation}

Here, $K_{p, x}, K_{p, y}>0$ are the user-defined proportional gains of the controller. 

\textit{Movement Towards Global Goal Direction:} In the last step, we align our heading direction (angle associated with Y axis) to orient ourselves towards the global goal direction using a proportional integrative derivative controller. After aligning with the goal direction, we move forward  (in the $Z$ direction)  for $\tau$ seconds using an open-loop positional command like before using a velocity controller. 

Finally, all the aforementioned steps which we call an iteration are repeated until we have reached the goal. 

Furthermore, the pre-chosen exploratory motion is generally much smaller in magnitude compared to the movement toward the free space. This formulation ensures that we do not spend too much time exploring and the overall path length is still relatively short as compared to having complete knowledge of depth. This is similar to the idea used in \cite{sanket2018gapflyt, bouwmeester2023nanoflownet}.

\subsection{High Priority Dynamic Obstacle Dodging}

Often in the wild, robots will encounter dynamic obstacles such as falling leaves, branches, or rocks in a disaster scenario. While navigating through a cluttered static scene, one has to prioritize dodging dynamic obstacles to avoid a severe catastrophe. To this end, our dynamic obstacle detection and control algorithm are always running in parallel to the static obstacle avoidance methodology as a separate thread. We detect dynamic obstacles by looking for a large magnitude of optical flow and uncertainty for consecutively two frames. Whenever a dynamic obstacle is detected, an interrupt is raised to invoke a dodging maneuver using an acceleration command in the direction as given in \cite{sanket2020evdodgenet}. The choice for the acceleration command is to minimize the latency and utilize the maximum agility of the robot.

\section{Experiments}
\label{sec:Expts}
The robot used in the experiments is a custom-built platform called Corgi210, and has a 210$mm$ diagonal wheelbase. 
All the lower-level control algorithms are run on the Holybro Pixhawk 32 V6 Flight Controller using ArduCopter version 4.5.0-Dev. The ArkFlow optical flow sensor is used for enabling Loiter. All the higher-level perception, decision making and control commands are computed onboard the NVIDIA Jetson Orin Nano running Ubuntu 20.04.6 LTS with JetPack 5.1.2 and TensorFlow 2.12.0. The higher-level control commands are sent to the flight controller using Guided mode. The perception stack takes input from the onboard Arducam OV9281 120fps global shutter camera at a resolution of $640 \times 480 px.$  




\subsection{Experimental Results}
\label{subsec:results}


\subsubsection{Evaluation Metrics}

We adapt \textit{Success Rate}, \textit{Path Length Increase} over metric depth, and \textit{Run Time} evaluation metrics from \cite{sanket2023ajna}. We perform exhaustive quantitative evaluation both in the real world and in simulation environments under various scenarios.








\subsubsection{Real World Experiments}
We test our \textit{MinNav} framework in the Washburn flying space which is a netted facility of dimension $11 \times 4.5 \times 3.65 m$. We construct four different scenarios with a variety of static and dynamic obstacles and an unknown-shaped gap (Fig. \ref{fig:experiments}). The static obstacles are made of cuboidal cardboard boxes of sizes ranging from $1.15 - 1.28m$ with rock and moss textures stuck on them. The dynamic obstacles are made of an empty milk jug and a cardboard box of sizes ranging from $0.25 - 0.30 m$ with the same textures as the static obstacles. These dynamic obstacles are swung (at a maximum speed of 3$ms^{-1}$) using a thread towards the quadrotor. We utilize two arbitrarily shaped gaps that are made of foamcore with stone textured paper on it. The gap has an antipodal minimum and a maximum distance of $0.5m$ and $0.85m$ respectively giving us an average minimum clearance of $0.17m$. 

In the four scenarios, the position and orientation of the static obstacles, gap shape and direction and velocity of the dynamic obstacles are randomized. Furthermore, we also randomize the ordering of the scene components (gap, static obstacles and dynamic obstacles). Specifically, \textbf{Scenario 1:} Randomized Static Obstacles $\rightarrow$ Gap Type 1 $\rightarrow$ Box dynamic obstacle, \textbf{Scenario 2:} Gap Type 1 $\rightarrow$ Randomized Static Obstacles $\rightarrow$ Milk carton dynamic obstacle, \textbf{Scenario 3:} Gap Type 2 $\rightarrow$ Randomized Static Obstacles $\rightarrow$ Box dynamic obstacle, \textbf{Scenario 4:} Randomized Static Obstacles  $\rightarrow$ Gap Type 2$\rightarrow$ Milk carton dynamic obstacle. 

We performed 20 trials on each scenario. Our method successfully could navigate through complex scenarios 56/80 times leading to an average success rate of $70\%$. Although this is not sufficient for ready deployment into the wild, we believe it is a first step in the direction of minimal autonomy in complex scenes in the wild. Furthermore, as the robot becomes smaller, its collision probability decreases because a reduced physical scale reduces its spatial footprint and improves operation in cluttered environments \cite{yu2023avoidbench}. This trend also motivates the development of insect-inspired cognitive systems, such as those inspired by bees \cite{chittka2023mind}.

\textbf{Discussion:} A majority of our failure cases are in scenarios when we are trying to dodge dynamic obstacles, as we hit the nets on the side of the robot due to a lack of omnidirectional sensing. Although, we did not explore better data generation regimes as described in \cite{singh2023worldgen, sun2021autoflow}, we believe they can significantly improve the optical flow quality and would be a great direction for future work. We also note that our latency for dynamic obstacle detection is about 100$ms$ including our processing pipeline, which is fairly large for fast and close obstacles. To this end, we plan to explore TensorRT accelerations as a future direction along with better software engineering in C++ to lower latency. Lastly, we aim to accelerate our optical flow predictions by utilizing methods from \cite{raju2024edgeflownet}, which will lower latency and improve dynamic obstacle response further. 

\subsection{Simulation Experiments}
\label{subsec:sim}


For a comprehensive quantitative evaluation of our approach, we designed a custom simulation environment in Blender$^{\text{\textregistered}}$ based on \cite{sanket2020evdodgenet, parameshwara20210}. The static components of the environment are represented by randomized trees depicting the challenges posed by natural obstacles. Dynamic obstacles were introduced in the form of randomly thrown soccer balls. The arbitrary-shaped gap was constructed using a planar surface with a randomized rock texture on it. The floor is simulated with a planar texture with bumps on it to mimic a real-world outdoor floor. We further randomize the ordering and location of the static objects and gaps and velocity + location of the dynamic obstacles. This deliberate variability in our generated scenes allowed us to conduct a comprehensive evaluation of \textit{MinNav's} robustness under different conditions. Upon acceptance of this publication, we will release these scenes and code with an open-source license to accelerate further research.


\begin{figure}
    \centering
    \includegraphics[width=0.95\linewidth]{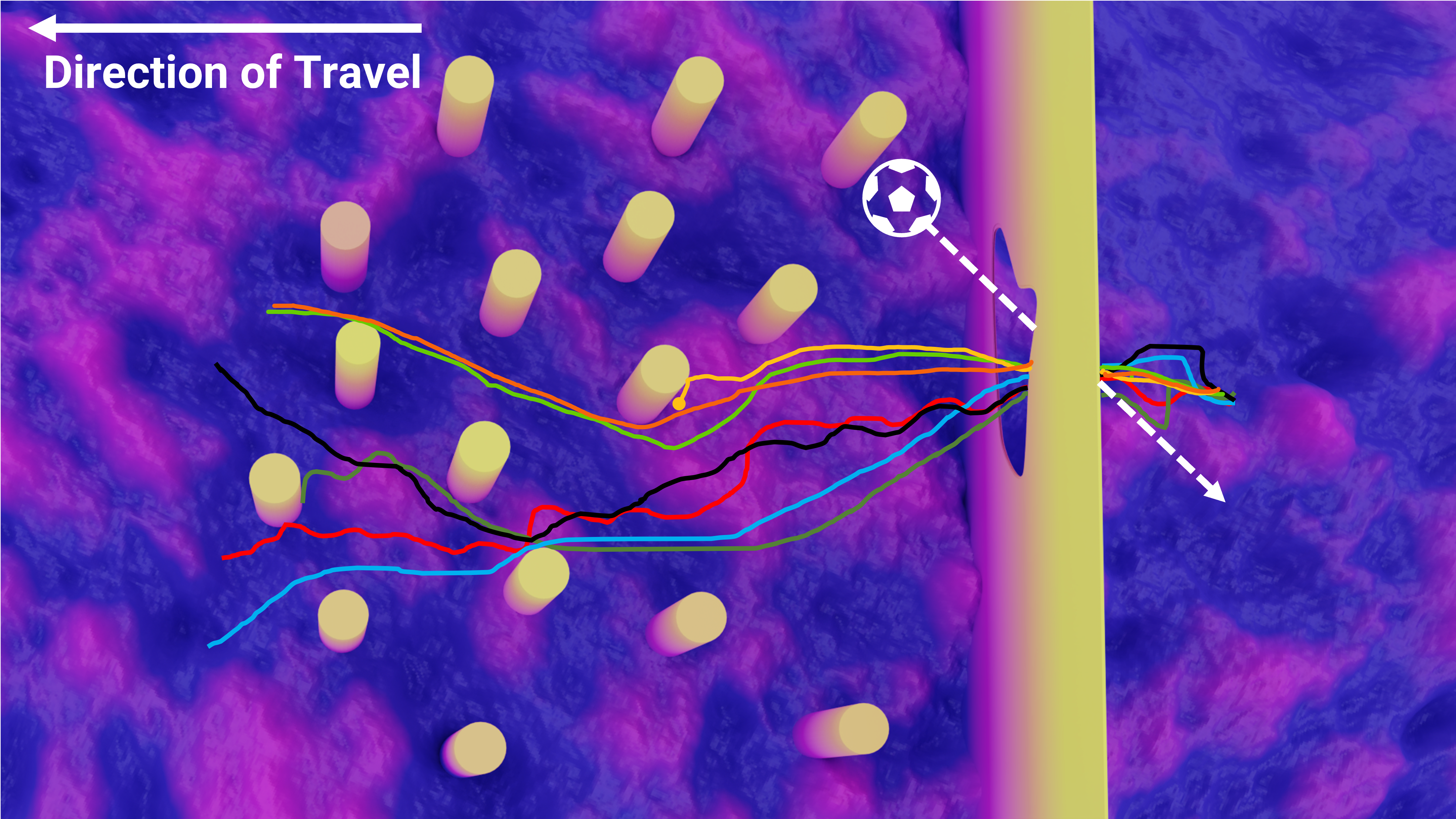}
    \caption{Comparison of various navigation methods: 
    \textcolor[RGB]{255, 140, 0}{Ground truth depth}, \textcolor[RGB]{102, 204, 0}{MiDaS-v3.1},  \textcolor[RGB]{255, 193, 20}{MiDaS-v2.1S}, \textcolor[RGB]{0, 176, 240}{RAFT without exploration}, \textbf{RAFT with exploration}, \textcolor[RGB]{84, 130, 53}{\textit{MinNav} without exploration}, \textcolor[RGB]{255, 0, 0}{\textit{MinNav} (Ours) with exploration}. The dashed white line shows the dynamic obstacle path. See Table \ref{tab:quantitativeval} for quantitative evaluation.
    \label{fig:SimEval}}
\end{figure}

We test our approach in 100 randomly simulated scenarios across various metrics and compare it with state-of-the-art approaches. In particular, we evaluate performance when the robot had access to different amounts of information, i.e., metric depth to relative monocular depth to optical flow (implicit function of metric depth).

For depth-based navigation, we use a simple strategy wherein we perform a potential field-based obstacle avoidance in a receding horizon manner to move towards the closest free space as described in \cite{sanket2021morpheyes}. This has the highest amount of information on the robot and results in the shortest path length, hence we measure all other path lengths as an increase (inefficiency) over the depth-based method. For the relative monocular depth method, we utilize a similar strategy as before but the robot parameters have to be more conservative due to a lack of environment knowledge. Finally, for optical flow-based methods, we utilize the \textit{MinNav} control algorithm on RAFT optical flow as well as our network. Furthermore, we also test our approach with and without exploratory motion to show the utility of active vision under a dearth of high-quality computation and sensing. The results are tabulated in Table \ref{tab:quantitativeval} and an example path output is shown in Fig. \ref{fig:SimEval}.

\textbf{Discussion:} As one would expect metric depth-based methods such as MorphEyes perform the best overall but require careful tuning of depth threshold parameters for dynamic obstacles which are obvious due to the lack of explicit motion information and can be the go-to choice for navigation if a depth sensor can be placed onboard the robot when coupled with motion information for dynamic obstacles (such as scene flow). Similarly, relative monocular depth-based methods are performant but often require a much higher amount of computation to predict high-fidelity depth. This is very evident from the high performance of the MiDAS-v3.1 model \cite{birkl2023midas, Ranftl2021} which is almost 2 factors of magnitude more expensive to compute than our model. Furthermore, it is important to observe that distilling a monocular depth model is not enough due to a massive loss in depth quality which fails to dodge obstacles as seen in the MiDAS-2.1-S model. This model not only fails to detect dynamic obstacles well due to their small size but also predicts inaccurate depth for large static obstacles resulting in a low success rate. Although optical flow-based methods require more care to be taken when dodging static obstacles, they excel at dynamic obstacles very well due to the virtue of their design. It is remarkable to observe that optical flow models although not more accurate than depth-based models, come close in overall success rate and have a massive potential to be used as a backup or sanity check algorithm for depth model failures. Furthermore, when the optical flow models become smaller, their accuracy reduces but the navigational success rate can be enhanced by adding an activeness component to the planner as seen in our work. Furthermore, it is noteworthy to see that \textit{MinNav} performs almost on par with RAFT in terms of success rate while being 9$\times$ faster embracing the active nature of the robot. We remark that smaller robots with a dearth of high-quality sensing and computation can leverage this methodology to enable successful navigation in the wild.

\begin{table}[t!]
    \caption{Quantitative Evaluation for Simulation Experiments.}
    \resizebox{\columnwidth}{!}
    {\label{tab:quantitativeval}
    \begin{tabular}{lccccc}
    \toprule
    \multirow{2}{*}{Method} & SR   \multirow{2}{*}{$\,\,\uparrow$} & Path Length \multirow{2}{*}{$\downarrow$} & Run Time$^\dagger$\multirow{2}{*}{$\downarrow$} & FLOPs \multirow{2}{*}{$\downarrow$}   & Num. Params \multirow{2}{*}{$\downarrow$} \\
        & ($\%$)  &  Inc. ($\%$) & (ms) & (G)  & (M) \\
    \midrule
      MorphEyes\cite{sanket2021morpheyes} & 100  & --  & 9.8  &  --   &  --  \\     
      \hline
      MiDAS-v2.1S\cite{Ranftl2022}      & 38 &  17.16 &  99.8 & 43.7   &  21  \\          
      MiDAS-v3.1\cite{birkl2023midas}     & \textbf{88} & \textbf{4.95} &  2915.2 & 1052.9 & 344.6 \\ 
      RAFT$^*$\cite{teed2020raft}       & 78  &  36.52 &  615.3  &  	211.01   & 5.2   \\ 
      RAFT\cite{teed2020raft}          &  84 & 32.91 &  615.3   &  	211.01   & 5.2   \\ 
      \hline
      \textit{MinNav}$^*$ \textit{(Ours)} & 57 & 57.84&  \textbf{68.2}  & \textbf{27.14}    & \textbf{2.8} \\ 
      \textit{MinNav} \textit{(Ours)} & 82 &52.49 &  \textbf{68.2}  & \textbf{27.14}    & \textbf{2.8} \\     
     \bottomrule
    \end{tabular}}
    \tiny{$^*$Without Exploration. $^\dagger$On NVIDIA Jetson Orin Nano at an input size  of $1 \times 640 \times 480 \times 6 px.$}
\end{table}

\section{Conclusions}
\label{sec:Conc}
We presented an active vision-based solution to navigate through a scene containing static and dynamic obstacles and unknown shaped gaps without the prior knowledge of scene components, their location and their ordering. We utilize exploratory motion on the quadrotor to alleviate common problems encountered with optical flow-based navigation. We show comparable results to depth-based navigation in both the real world and simulation. We believe such a strategy can help advance the autonomy of tiny aerial robots further. As a parting thought, utilizing optical flow and uncertainty to dodge thin wire-like obstacles is a promising direction we will explore further to enhance the navigational capabilities of small aerial robots that will bring large-scale deployment into the wild one step further.

\bibliographystyle{unsrt}
\bibliography{Ref}

@article{burnett2020wind,
  title={Wind and obstacle motion affect honeybee flight strategies in cluttered environments},
  author={Burnett, Nicholas P and Badger, Marc A and Combes, Stacey A},
  journal={Journal of Experimental Biology},
  volume={223},
  number={14},
  pages={jeb222471},
  year={2020},
  publisher={The Company of Biologists Ltd}
}

@article{zhang2017online,
  title={Online lidar and vision based ego-motion estimation and mapping},
  author={Zhang, Ji},
  journal={PhD thesis},
  year={2017},
  publisher={Carnegie Mellon University}
}

@phdthesis{sanket2021active,
  title={Active Vision Based Embodied-AI Design for Nano-UAV Autonomy},
  author={Sanket, Nitin Jagannatha},
  year={2021},
  school={University of Maryland, College Park}
}

@article{mazinani2023design,
  title={Design and analysis of an aerial pollination system for walnut trees},
  author={Mazinani, Mozhdeh and Zarafshan, Payam and Dehghani, Mohammad and Vahdati, Kourosh and Etezadi, Hamed},
  journal={Biosystems Engineering},
  volume={225},
  pages={83--98},
  year={2023},
  publisher={Elsevier}
}

@inproceedings{mohta2016quadcloud,
  title={Quadcloud: a rapid response force with quadrotor teams},
  author={Mohta, Kartik and Turpin, Matthew and Kushleyev, Alex and Mellinger, Daniel and Michael, Nathan and Kumar, Vijay},
  booktitle={Experimental Robotics: The 14th International Symposium on Experimental Robotics},
  pages={577--590},
  year={2016},
  organization={Springer}
}

@article{delmerico2019current,
  title={The current state and future outlook of rescue robotics},
  author={Delmerico, Jeffrey and Mintchev, Stefano and Giusti, Alessandro and Gromov, Boris and Melo, Kamilo and Horvat, Tomislav and Cadena, Cesar and Hutter, Marco and Ijspeert, Auke and Floreano, Dario and others},
  journal={Journal of Field Robotics},
  volume={36},
  number={7},
  pages={1171--1191},
  year={2019},
  publisher={Wiley Online Library}
}

@inproceedings{Inspection,
  title={{Inspection of penstocks and featureless tunnel-like environments using micro UAVs}},
  author={{\"O}zaslan, T. and  others},
  booktitle={Field and Service Robotics},
  pages={123--136},
  year={2015},
  organization={Springer}
}

@article{SearchAndRescue,
  title={Collaborative mapping of an earthquake-damaged building via ground and aerial robots},
  author={Michael, N. and others},
  journal={Journal of Field Robotics},
  volume={29},
  number={5},
  pages={832--841},
  year={2012},
  publisher={Wiley Online Library}
}

@article{souhila2007optical,
  title={Optical flow based robot obstacle avoidance},
  author={Souhila, Kahlouche and Karim, Achour},
  journal={International Journal of Advanced Robotic Systems},
  volume={4},
  number={1},
  pages={2},
  year={2007},
  publisher={SAGE Publications Sage UK: London, England}
}

@inproceedings{duchon1995ecological,
  title={Ecological robotics: Controlling behavior with optical flow},
  author={Duchon, Andrew P and Warren, William H and Kaelbling, L Pack},
  booktitle={Proceedings of the seventeenth annual conference of the Cognitive Science Society},
  volume={17},
  pages={164},
  year={1995},
  organization={Psychology Press}
}

@article{horn1980determining,
  title={Determining optical flow},
  author={Horn, Berthold KP and Schunck, Brian G},
  year={1980}
}

@article{ActiveVision,
  title={Active vision},
  author={J. Aloimonos and others},
  journal={International journal of computer vision},
  volume={1},
  number={4},
  pages={333--356},
  year={1988},
  publisher={Springer}
}

@article{sanket2018gapflyt,
  title={Gapflyt: Active vision based minimalist structure-less gap detection for quadrotor flight},
  author={Sanket, Nitin J and Singh, Chahat Deep and Ganguly, Kanishka and Ferm{\"u}ller, Cornelia and Aloimonos, Yiannis},
  journal={IEEE Robotics and Automation Letters},
  volume={3},
  number={4},
  pages={2799--2806},
  year={2018},
  publisher={IEEE}
}

@article{sanket2023ajna,
  title={Ajna: Generalized deep uncertainty for minimal perception on parsimonious robots},
  author={Sanket, Nitin J and Singh, Chahat Deep and Ferm{\"u}ller, Cornelia and Aloimonos, Yiannis},
  journal={Science Robotics},
  volume={8},
  number={81},
  pages={eadd5139},
  year={2023},
  publisher={American Association for the Advancement of Science}
}

@inproceedings{bouwmeester2023nanoflownet,
  title={Nanoflownet: Real-time dense optical flow on a nano quadcopter},
  author={Bouwmeester, Rik J and Paredes-Vall{\'e}s, Federico and De Croon, Guido CHE},
  booktitle={2023 IEEE International Conference on Robotics and Automation (ICRA)},
  pages={1996--2003},
  year={2023},
  organization={IEEE}
}

@inproceedings{sanket2020evdodgenet,
  title={Evdodgenet: Deep dynamic obstacle dodging with event cameras},
  author={Sanket, Nitin J and Parameshwara, Chethan M and Singh, Chahat Deep and Kuruttukulam, Ashwin V and Ferm{\"u}ller, Cornelia and Scaramuzza, Davide and Aloimonos, Yiannis},
  booktitle={2020 IEEE International Conference on Robotics and Automation (ICRA)},
  pages={10651--10657},
  year={2020},
  organization={IEEE}
}

@inproceedings{dupeyroux2021neuromorphic,
  title={Neuromorphic control for optic-flow-based landing of MAVs using the Loihi processor},
  author={Dupeyroux, Julien and Hagenaars, Jesse J and Paredes-Vall{\'e}s, Federico and de Croon, Guido CHE},
  booktitle={2021 IEEE International Conference on Robotics and Automation (ICRA)},
  pages={96--102},
  year={2021},
  organization={IEEE}
}

@inproceedings{stevens2018vision,
  title={Vision based forward sensitive reactive control for a quadrotor VTOL},
  author={Stevens, Jean-Luc and Mahony, Robert},
  booktitle={2018 IEEE/RSJ International Conference on Intelligent Robots and Systems (IROS)},
  pages={5232--5238},
  year={2018},
  organization={IEEE}
}

@article{shah2021traditional,
  title={Traditional and modern strategies for optical flow: an investigation},
  author={Shah, Syed Tafseer Haider and Xuezhi, Xiang},
  journal={SN Applied Sciences},
  volume={3},
  pages={1--14},
  year={2021},
  publisher={Springer}
}

@inproceedings{sanket2021morpheyes,
  title={Morpheyes: Variable baseline stereo for quadrotor navigation},
  author={Sanket, Nitin J and Singh, Chahat Deep and Asthana, Varun and Ferm{\"u}ller, Cornelia and Aloimonos, Yiannis},
  booktitle={2021 IEEE International Conference on Robotics and Automation (ICRA)},
  pages={413--419},
  year={2021},
  organization={IEEE}
}

@ARTICLE {Ranftl2022,
    author  = "Ren\'{e} Ranftl and Katrin Lasinger and David Hafner and Konrad Schindler and Vladlen Koltun",
    title   = "Towards Robust Monocular Depth Estimation: Mixing Datasets for Zero-Shot Cross-Dataset Transfer",
    journal = "IEEE Transactions on Pattern Analysis and Machine Intelligence",
    year    = "2022",
    volume  = "44",
    number  = "3"
}

@article{Ranftl2021,
	author    = {Ren\'{e} Ranftl and Alexey Bochkovskiy and Vladlen Koltun},
	title     = {Vision Transformers for Dense Prediction},
	journal   = {ICCV},
	year      = {2021},
}

@article{birkl2023midas,
      title={MiDaS v3.1 -- A Model Zoo for Robust Monocular Relative Depth Estimation},
      author={Reiner Birkl and Diana Wofk and Matthias M{\"u}ller},
      journal={arXiv preprint arXiv:2307.14460},
      year={2023}
}

@inproceedings{teed2020raft,
  title={Raft: Recurrent all-pairs field transforms for optical flow},
  author={Teed, Zachary and Deng, Jia},
  booktitle={Computer Vision--ECCV 2020: 16th European Conference, Glasgow, UK, August 23--28, 2020, Proceedings, Part II 16},
  pages={402--419},
  year={2020},
  organization={Springer}
}

@article{gao2019flying,
  title={Flying on point clouds: Online trajectory generation and autonomous navigation for quadrotors in cluttered environments},
  author={Gao, Fei and Wu, William and Gao, Wenliang and Shen, Shaojie},
  journal={Journal of Field Robotics},
  volume={36},
  number={4},
  pages={710--733},
  year={2019},
  publisher={Wiley Online Library}
}

@article{zhou2022swarm,
  title={Swarm of micro flying robots in the wild},
  author={Zhou, Xin and Wen, Xiangyong and Wang, Zhepei and Gao, Yuman and Li, Haojia and Wang, Qianhao and Yang, Tiankai and Lu, Haojian and Cao, Yanjun and Xu, Chao and others},
  journal={Science Robotics},
  volume={7},
  number={66},
  pages={eabm5954},
  year={2022},
  publisher={American Association for the Advancement of Science}
}

@inproceedings{liu2016high,
  title={High speed navigation for quadrotors with limited onboard sensing},
  author={Liu, Sikang and Watterson, Michael and Tang, Sarah and Kumar, Vijay},
  booktitle={2016 IEEE international conference on robotics and automation (ICRA)},
  pages={1484--1491},
  year={2016},
  organization={IEEE}
}

@article{hanover2023autonomous,
  title={Autonomous drone racing: A survey},
  author={Hanover, Drew and Loquercio, Antonio and Bauersfeld, Leonard and Romero, Angel and Penicka, Robert and Song, Yunlong and Cioffi, Giovanni and Kaufmann, Elia and Scaramuzza, Davide},
  journal={arXiv e-prints, pp. arXiv--2301},
  year={2023}
}

@article{song2023reaching,
  title={Reaching the limit in autonomous racing: Optimal control versus reinforcement learning},
  author={Song, Yunlong and Romero, Angel and M{\"u}ller, Matthias and Koltun, Vladlen and Scaramuzza, Davide},
  journal={Science Robotics},
  volume={8},
  number={82},
  pages={eadg1462},
  year={2023},
  publisher={American Association for the Advancement of Science}
}

@article{kaufmann2023champion,
  title={Champion-level drone racing using deep reinforcement learning},
  author={Kaufmann, Elia and Bauersfeld, Leonard and Loquercio, Antonio and M{\"u}ller, Matthias and Koltun, Vladlen and Scaramuzza, Davide},
  journal={Nature},
  volume={620},
  number={7976},
  pages={982--987},
  year={2023},
  publisher={Nature Publishing Group UK London}
}

@article{eschmann2023learning,
  title={Learning to Fly in Seconds},
  author={Eschmann, Jonas and Albani, Dario and Loianno, Giuseppe},
  journal={arXiv preprint arXiv:2311.13081},
  year={2023}
}

@article{ferede2024end,
  title={End-to-end neural network based optimal quadcopter control},
  author={Ferede, Robin and de Croon, Guido and De Wagter, Christophe and Izzo, Dario},
  journal={Robotics and Autonomous Systems},
  volume={172},
  pages={104588},
  year={2024},
  publisher={Elsevier}
}

@article{izzo2012landing,
  title={Landing with time-to-contact and ventral optic flow estimates},
  author={Izzo, Dario and De Croon, Guido},
  journal={Journal of Guidance, Control, and Dynamics},
  volume={35},
  number={4},
  pages={1362--1367},
  year={2012}
}

@book{chittka2023mind,
  title={The mind of a bee},
  author={Chittka, Lars},
  year={2023},
  publisher={Princeton University Press}
}

@article{ravi2019gap,
  title={Gap perception in bumblebees},
  author={Ravi, Sridhar and Bertrand, Olivier and Siesenop, Tim and Manz, Lea-Sophie and Doussot, Charlotte and Fisher, Alex and Egelhaaf, Martin},
  journal={Journal of Experimental Biology},
  volume={222},
  number={2},
  pages={jeb184135},
  year={2019},
  publisher={The Company of Biologists Ltd}
}

@InProceedings{DFIB15,
  author    = "A. Dosovitskiy and P. Fischer and E. Ilg and P. H{\"a}usser and C. Haz{\i}rba{\c{s}} and V. Golkov and P. v.d. Smagt and D. Cremers and T. Brox",
  title     = "FlowNet: Learning Optical Flow with Convolutional Networks",
  booktitle = "IEEE International Conference on Computer Vision (ICCV)",
  month     = " ",
  year      = "2015",
  url       = "http://lmb.informatik.uni-freiburg.de/Publications/2015/DFIB15"
}

@InProceedings{ISKB18,
  author    = "E. Ilg and T. Saikia and M. Keuper and T. Brox",
  title     = "Occlusions, Motion and Depth Boundaries with a Generic Network for Disparity, Optical Flow or Scene Flow Estimation",
  booktitle = "European Conference on Computer Vision (ECCV)",
  month     = " ",
  year      = "2018",
  url       = "http://lmb.informatik.uni-freiburg.de/Publications/2018/ISKB18"
}

@InProceedings{MIFDB16,
  author    = "N. Mayer and E. Ilg and P. H{\"a}usser and P. Fischer and D. Cremers and A. Dosovitskiy and T. Brox",
  title     = "A Large Dataset to Train Convolutional Networks for Disparity, Optical Flow, and Scene Flow Estimation",
  booktitle = "IEEE International Conference on Computer Vision and Pattern Recognition (CVPR)",
  year      = "2016",
  note      = "arXiv:1512.02134",
  url       = "http://lmb.informatik.uni-freiburg.de/Publications/2016/MIFDB16"
}

@article{sanket2021evpropnet,
  title={EVPropNet: Detecting drones by finding propellers for mid-air landing and following},
  author={Sanket, Nitin J and Singh, Chahat Deep and Parameshwara, Chethan M and Ferm{\"u}ller, Cornelia and de Croon, Guido CHE and Aloimonos, Yiannis},
  journal={arXiv preprint arXiv:2106.15045},
  year={2021}
}

@inproceedings{singh2021nudgeseg,
  title={Nudgeseg: Zero-shot object segmentation by repeated physical interaction},
  author={Singh, Chahat Deep and Sanket, Nitin J and Parameshwara, Chethan M and Ferm{\"u}ller, Cornelia and Aloimonos, Yiannis},
  booktitle={2021 IEEE/RSJ International Conference on Intelligent Robots and Systems (IROS)},
  pages={2714--2712},
  year={2021},
  organization={IEEE}
}

@article{BajcsyActive,
  title={Revisiting active perception},
  author={Ruzena Bajcsy and others},
  journal={Autonomous Robots},
  pages={1--20},
  year={2017},
  publisher={Springer}
}

@article{ravi2022bumblebees,
  title={Bumblebees display characteristics of active vision during robust obstacle avoidance flight},
  author={Ravi, Sridhar and Siesenop, Tim and Bertrand, Olivier J and Li, Liang and Doussot, Charlotte and Fisher, Alex and Warren, William H and Egelhaaf, Martin},
  journal={Journal of Experimental Biology},
  volume={225},
  number={4},
  pages={jeb243021},
  year={2022},
  publisher={The Company of Biologists Ltd}
}

@inproceedings{parameshwara20210,
  title={0-mms: Zero-shot multi-motion segmentation with a monocular event camera},
  author={Parameshwara, Chethan M and Sanket, Nitin J and Singh, Chahat Deep and Ferm{\"u}ller, Cornelia and Aloimonos, Yiannis},
  booktitle={2021 IEEE International Conference on Robotics and Automation (ICRA)},
  pages={9594--9600},
  year={2021},
  organization={IEEE}
}

@inproceedings{singh2023worldgen,
  title={WorldGen: A Large Scale Generative Simulator},
  author={Singh, Chahat Deep and Kumari, Riya and Ferm{\"u}ller, Cornelia and Sanket, Nitin J and Aloimonos, Yiannis},
  booktitle={2023 IEEE International Conference on Robotics and Automation (ICRA)},
  pages={9147--9154},
  year={2023},
  organization={IEEE}
}

@inproceedings{sun2021autoflow,
  title={Autoflow: Learning a better training set for optical flow},
  author={Sun, Deqing and Vlasic, Daniel and Herrmann, Charles and Jampani, Varun and Krainin, Michael and Chang, Huiwen and Zabih, Ramin and Freeman, William T and Liu, Ce},
  booktitle={Proceedings of the IEEE/CVF Conference on Computer Vision and Pattern Recognition},
  pages={10093--10102},
  year={2021}
}

@article{gapflyt, 
author={N. Sanket and others},
journal={IEEE Robotics and Automation Letters}, 
title={{GapFlyt}: Active Vision Based Minimalist Structure-Less Gap Detection For Quadrotor Flight}, 
year={2018}, 
volume={3}, 
number={4}, 
pages={2799-2806}, 
ISSN={}, 
month={Oct},}

@inproceedings{zhong2019unsupervised,
  title={Unsupervised deep epipolar flow for stationary or dynamic scenes},
  author={Zhong, Yiran and Ji, Pan and Wang, Jianyuan and Dai, Yuchao and Li, Hongdong},
  booktitle={Proceedings of the IEEE/CVF conference on computer vision and pattern recognition},
  pages={12095--12104},
  year={2019}
}

@inproceedings{ranjan2019competitive,
  title={Competitive collaboration: Joint unsupervised learning of depth, camera motion, optical flow and motion segmentation},
  author={Ranjan, Anurag and Jampani, Varun and Balles, Lukas and Kim, Kihwan and Sun, Deqing and Wulff, Jonas and Black, Michael J},
  booktitle={Proceedings of the IEEE/CVF conference on computer vision and pattern recognition},
  pages={12240--12249},
  year={2019}
}

@article{raju2024edgeflownet,
  title={EdgeFlowNet: 100FPS@ 1W Dense Optical Flow for Tiny Mobile Robots},
  author={Raju, Sai Ramana Kiran Pinnama and Singh, Rishabh and Velmurugan, Manoj and Sanket, Nitin J},
  journal={IEEE Robotics and Automation Letters},
  year={2024},
  publisher={IEEE}
}

@article{sanket2021prgflow,
  title={PRGFlow: Unified SWAP-aware deep global optical flow for aerial robot navigation},
  author={Sanket, Nitin J and Singh, Chahat Deep and Ferm{\"u}ller, Cornelia and Aloimonos, Yiannis},
  journal={Electronics Letters},
  volume={57},
  number={16},
  pages={614--617},
  year={2021},
  publisher={Wiley Online Library}
}

@article{ren2025safety,
  title={Safety-assured high-speed navigation for MAVs},
  author={Ren, Yunfan and Zhu, Fangcheng and Lu, Guozheng and Cai, Yixi and Yin, Longji and Kong, Fanze and Lin, Jiarong and Chen, Nan and Zhang, Fu},
  journal={Science Robotics},
  volume={10},
  number={98},
  pages={eado6187},
  year={2025},
  publisher={American Association for the Advancement of Science}
}

@inproceedings{kulkarni2023semantically,
  title={Semantically-enhanced deep collision prediction for autonomous navigation using aerial robots},
  author={Kulkarni, Mihir and Nguyen, Huan and Alexis, Kostas},
  booktitle={2023 IEEE/RSJ International Conference on Intelligent Robots and Systems (IROS)},
  pages={3056--3063},
  year={2023},
  organization={IEEE}
}

@article{yu2023avoidbench,
  title={Avoidbench: A high-fidelity vision-based obstacle avoidance benchmarking suite for multi-rotors},
  author={Yu, Hang and de Croon, Guido CH and De Wagter, Christophe},
  journal={arXiv preprint arXiv:2301.07430},
  year={2023}
}

@misc{ranjan2016opticalflowestimationusing,
      title={Optical Flow Estimation using a Spatial Pyramid Network}, 
      year={2016},
      eprint={1611.00850},
      archivePrefix={arXiv},
      primaryClass={cs.CV},
      url={https://arxiv.org/abs/1611.00850}, 
}

\end{document}